# Generalized Two Color Map Theorem -- Complete Theorem of Robust Gait Plan for a Tilt-rotor

Zhe Shen[1], Yudong Ma[2] and Takeshi Tsuchiya[2]

*Abstract—* Gait plan is a procedure that is typically applied on the ground robots, e.g., quadrupedal robots; the tilt-rotor, a novel type of quadrotor with eight inputs, is not one of them. While controlling the tilt-rotor relying on feedback linearization, the tilting angles (inputs) are expected to change over-intensively, which may not be expected in the application. To help suppress the intensive change in the tilting angles, a gait plan procedure is introduced to the tilt-rotor before feedback linearization. The tilting angles are specified with time in advance by users rather than be given by the control rule. However, based on this scenario, the decoupling matrix in feedback linearization can be singular for some attitudes, combinations of roll angle and pitch angle. It hinders the further application of the feedback linearization. With this concern, Two Color Map Theorem is established to maximize the acceptable attitude region, where the combinations of roll and pitch will give an invertible decoupling matrix. That theorem, however, over-restricts the choice of the tilting angles, which can rule out some feasible robust gaits. This paper gives the generalized Two Color Map Theorem; all the robust gaits can be found based on this generalized theorem. The robustness of three gaits that satisfy this Generalized Two Color Map Theorem (while violating Two Color Map Theorem) are analyzed. The results show that Generalized Two Color Map Theorem completes the search for the robust gaits for a tilt-rotor.

## I. INTRODUCTION

It has been an exact decade since the first time Ryll's tilt-rotor was put forward and stabilized [1]. Comparing with the conventional quadrotor, this UAV attracts attentions since its unique capability of generating the lateral forces. As shown in Fig. 1 [2], [3], the directions of the thrusts are able to be adjusted by changing the tilting angles, $\alpha_1, \alpha_2, \alpha_3, \alpha_4$, during the flight.

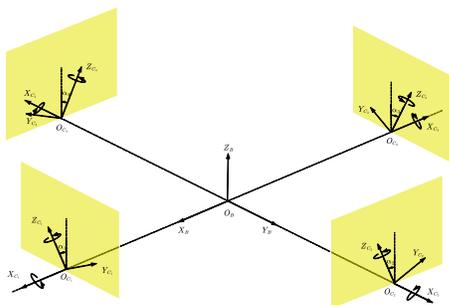

Fig. 1. The structure of Ryll's tilt-rotor. The directions of the thrusts are adjustable on the yellow planes during the flight.

Augmented with four tilting angles, the number of inputs in the tilt-rotor increases to eight from four, four magnitudes of the thrusts. Attempted by the over-actuated property, all the degrees of freedom (six) were initially stabilized by feedback linearization [4], [5].

Besides feedback linearization, several linear and nonlinear control methods are also applied to stabilize the tilt-rotor. These controllers include LQR and PID [6], [7], optimal control [8], [9], backstepping and sliding mode control [10]–[12], adaptive control [13], [14], etc. All these control methods can be classified according to the number of inputs calculated by the control rule [15]. Most of these controllers set eight inputs in a united control rule. Thus, the number of inputs for these controllers is eight. While some of them only calculate six inputs [16], the rest inputs are received by synchronizing the tilting angles. The number of inputs for these controllers is six.

Despite the success in some tracking problem [17]–[19] for the tilt-rotor, the commonly used feedback linearization (dynamic inversion) technique may suffer from the so-called over-intensive change [20], [21] in the tilting angles. These tilting angles are required to change greatly within a short time step [4] or to change beyond one cycle [1]; the tilting angles are required to spin several rounds. For example, Fig. 2 is the tilting angle history in a hovering problem by feedback linearization for a tilt-rotor [2]. The tilting angles are supposed to change in high frequencies and on a large scale at the beginning.

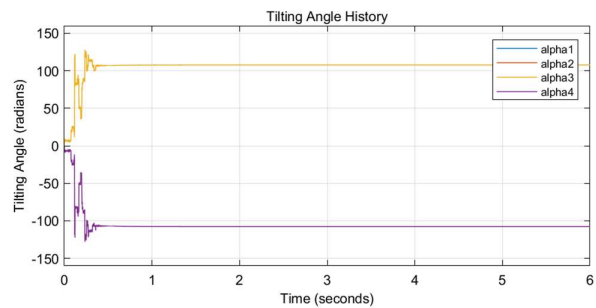

Fig. 2. The tilting angle history of a tilt-rotor while hovering.

These over-intensive changes in the tilting angles are resulted by the nature of feedback linearization. At each sampling time, the controller calculates the inverse of the dynamics, the result of which can be discontinuous, driving the tilting angles to change intensively. It is worth mentioning that this phenomenon is not unique in feedback linearization [22].

[1]Department of Aeronautics and Astronautics, The University of Tokyo, Tokyo 113-8654 Japan (+81-080-3709-1024; zheshen@g.ecc.u-tokyo.ac.jp).

[2]Department of Aeronautics and Astronautics, The University of Tokyo, Tokyo 113-8654 Japan.

To stabilize the tilt-rotor by feedback linearization while avoiding the over-intensive changes in the tilting angles, our previous research put forward a scenario where the tilting angles are planned in a procedure called gait plan in advance before the subsequent feedback linearization [2]. In the gait plan procedure, four tilting angles are specified with time by the user, which is parallel to the gait plan procedure in a ground robot. The subsequent feedback linearization and controller only assign the four magnitudes of the thrusts. Then, the calculated magnitudes of the thrusts and the tilting angles (gait) planned are the eight inputs to stabilize the tilt-rotor.

It seems that the gait plan is totally independent of the subsequent feedback linearization; the tilting angles are specified by the users, which is not intervened by the controller. Several animal-inspired gaits [23]–[25] are adopted and show their success in the reference-tracking problems. Note that the number of inputs calculated by the controller is four, based on this scenario, marking the birth of the new branch of the tilt-rotor controllers.

As for the subsequent feedback linearization after the gait plan, the decoupling matrix is required to be invertible; a singular decoupling matrix will cause the failure in calculating finite control signal [26]–[28]. The singularity of the decoupling matrix, in this case, is highly influenced by the tilting angles as well as the attitude, roll and pitch in specific. The tilt-rotor will not be allowed to maneuver to several roll and pitch angles [2]. We call these roll and pitch angles as unacceptable attitudes.

Clearly, for roll-pitch diagram fully occupied by the unacceptable attitudes, given the specific gait, feedback linearization is not able to control the tilt-rotor. On the other hand, a roll-pitch diagram without the occupation of the unacceptable attitudes gives more freedom for the feedback linearization to control the tilt-rotor; it leaves more acceptable attitudes for the feedback linearization. Generally, the larger the acceptable region of attitudes is, the more robust this gait is.

Our previous research [23] proved that the robustness can be increased by the scaling method. Scaling the gait enlarges the acceptable region of attitudes. However, this method partially sacrifices the lateral force. In alternative, Two Color Map Theorem [29], [30] is established to guide the gait plan in a general way, which gives more freedom to plan a desired gait. The limit of the Two Color Map Theorem is that the choice of $\alpha_1, \alpha_2$ is not arbitrary, which narrows the exploration of the robust gaits.

The main contribution of this paper is to advance Two Color Map Theorem. The generalized theorem, Generalized Two Color Map Theorem, is a complete theorem that includes all the continuous robust gaits. The robustness of several continuous robust gaits, which are ignored by Two Color Map Theorem but can be searched by Generalized Two Color Map Theorem, will be analyzed later in this paper.

The rest of this paper is organized as follows. Section II introduces the related works on Two Color Map Theorem. A generalized theorem, Generalized Two Color Map Theorem, is put forward in Section III. The robustness of three robust gaits satisfying Generalized Two Color Map Theorem is analyzed in Section IV. Finally, Section V addresses the conclusions and discussions.

II. RELATED WORK IN GAIT PLAN FOR A TILT-ROTOR

*A. Two Color Map Theorem*

A gait is the combination of four time-specified tilting angles, $\alpha_1(t), \alpha_2(t), \alpha_3(t), \alpha_4(t)$. Two Color Map Theorem guides the gait plan to create robust gaits, avoiding introducing the invertible decoupling matrix in large range in the subsequent feedback linearization in a large attitude region.

As the first step, users specify $\alpha_1(t), \alpha_2(t)$, continuously. In Two Color Map Theorem [29], $\alpha_1(t), \alpha_2(t)$ are required to vary while satisfying (1) ~ (2) for the time point $t_1$ where $\dot{\alpha}_1$ and $\dot{\alpha}_2$ exist.

$$\begin{vmatrix} \dot{\alpha}_1(t_1) & \dot{\alpha}_2(t_1) \\ 1 & 0 \end{vmatrix} = 0 \text{ or } \begin{vmatrix} \dot{\alpha}_1(t_1) & \dot{\alpha}_2(t_1) \\ 0 & 1 \end{vmatrix} = 0. \quad (1)$$

$$\|[\dot{\alpha}_1(t_1) \quad \dot{\alpha}_2(t_1)]\|_2 \neq 0. \quad (2)$$

In plain words, $(\alpha_1(t), \alpha_2(t))$ is required to move either horizontally or vertically on the $\alpha_1(t) - \alpha_2(t)$ diagram at any given time point. Users are allowed to specify a continuous $(\alpha_1(t), \alpha_2(t))$ freely based on this pattern.

After specifying $(\alpha_1(t), \alpha_2(t))$, $(\alpha_3(t), \alpha_4(t))$ are to be specified in the second step. Our previous research [29] proved that: To create a robust gait, there are only two candidates of $(\alpha_3, \alpha_4)$ corresponding to a given $(\alpha_1, \alpha_2)$, in general.

Further, observed from the $O\alpha_1\alpha_2\alpha_3$ coordinate system, these candidate $\alpha_3$ distribute on two planes. Also, observed from the $O\alpha_1\alpha_2\alpha_4$ coordinate system, these candidate $\alpha_4$ distribute on another two planes. We refer the distributions of these candidate $(\alpha_3, \alpha_4)$ to our previous publication [29].

Color the candidate $(\alpha_3, \alpha_4)$ on the first two corresponding planes in $O\alpha_1\alpha_2\alpha_3$ and $O\alpha_1\alpha_2\alpha_4$ red. Color the candidate $(\alpha_3, \alpha_4)$ on the other two corresponding planes in $O\alpha_1\alpha_2\alpha_3$ and $O\alpha_1\alpha_2\alpha_4$ blue. Then we receive the Two Color Map of $\alpha_1(t) - \alpha_2(t)$ diagram. A part of the Two Color Map is displayed in Fig. 3 [29], where the $(\alpha_1, \alpha_2)$ in half red and half blue represent that there are two candidate $(\alpha_3, \alpha_4)$. The user has to decide one. We refer the complete Two Color Map to the publication [30].

With this property, the selection of the candidate $(\alpha_3, \alpha_4)$ for each $(\alpha_1, \alpha_2)$ shall satisfy a rule to guarantee that the designed gait $(\alpha_1(t), \alpha_2(t), \alpha_3(t), \alpha_4(t))$ is continuous. This rule is called Two Color Map Theorem [29].

*Two Color Map Theorem*: The planned gait is continuous if the color selections of $(\alpha_1, \alpha_2)$ on the enclosed curve meet the following requirement: The adjacent $(\alpha_1, \alpha_2)$ are in the same color or do not violate the crossover rule.

*Proof*: See [29].

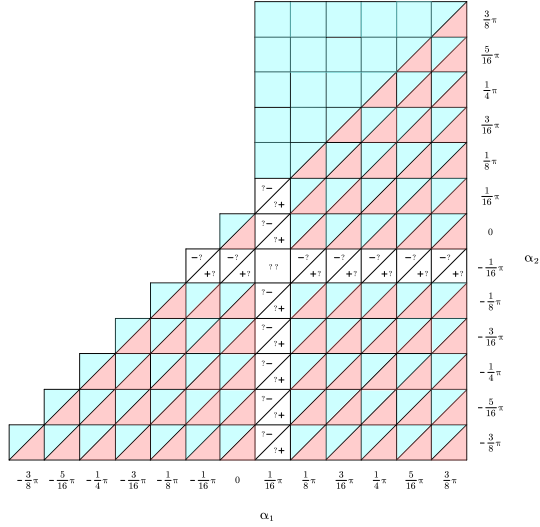

Fig. 3. A part of Two Color Map. The red $(\alpha_1, \alpha_2)$ correspond to the $(\alpha_3, \alpha_4)$ where all $\alpha_3$ are on the same plane and all $\alpha_4$ are on the same plane. The $(\alpha_1, \alpha_2)$ in half red and half blue represent that there are two candidate $(\alpha_3, \alpha_4)$. The user has to decide one.

The gaits planned on the Two Color Map will be robust gaits. While the robust gaits which satisfies Two Color Map Theorem in the meanwhile are the continuous robust gaits, which are expected in application.

### B. Surfaces in Two Color Map Theorem

The robustness of each gait is evaluated by the region of the unacceptable attitudes on the roll-pitch diagram, which will introduce the singular decoupling matrix in the feedback linearization [26], [31].

The determinant of the decoupling matrix is influenced by six variables, $\alpha_1$, $\alpha_2$, $\alpha_3$, $\alpha_4$, $\phi$ (roll), and $\theta$ (pitch). After specifying the gait with time, $(\alpha_1(t), \alpha_2(t), \alpha_3(t), \alpha_4(t))$, the combinations of $(\phi, \theta)$ which results zero determinants can be found on the $\phi - \theta$ diagram. These attitudes are not acceptable which hinder the application of feedback linearization.

In general, the further these curves are from the $(\phi, \theta) = (0,0)$, the more robust the gait is. For example, the curves in Fig. 4 and Fig. 5 are the unacceptable attitudes for the two gaits on the blue surface and the red surface, respectively. Obviously, the gait on the red surface (Fig. 5) is more robust. The details of these two gaits have been illustrated in our previous research [29].

One may notice that the robustness of these two gaits on two surfaces varies greatly although both gaits satisfy Two Color Map Theorem. This is because that Two Color Map Theorem only guarantees that the attitude region ***near*** $(\phi, \theta) = (0,0)$ introduces the invertible decoupling matrix [29]. Whereas the further region is not taken into consideration by this theorem.

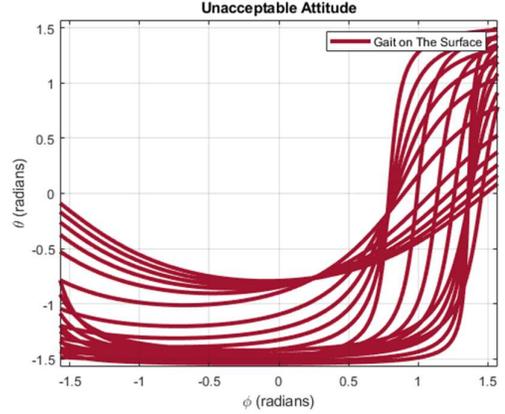

Fig. 4. The unacceptable attitudes for one gait on the blue surface. All these curves represent the attitudes that will introduce the singular decoupling matrix in feedback linearization. Note that these curves are relatively close to $(\phi, \theta) = (0,0)$.

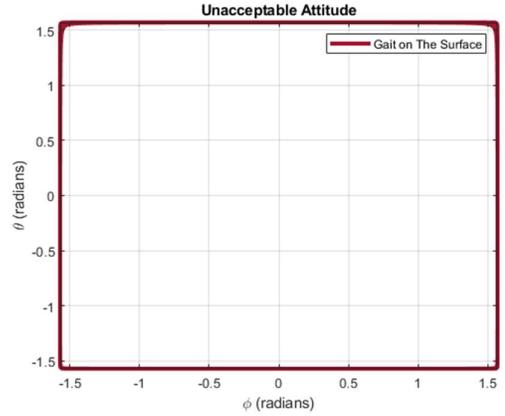

Fig. 5. The unacceptable attitudes for one gait on the red surface. All these curves represent the attitudes that will introduce the singular decoupling matrix in feedback linearization. Note that these curves are far from $(\phi, \theta) = (0,0)$

The underlying mechanism is that the coefficient of $\phi$ and $\theta$ ($R_\phi(\alpha_1, \alpha_2, \alpha_3, \alpha_4)$ and $R_\theta(\alpha_1, \alpha_2, \alpha_3, \alpha_4)$ in the Ref. [29]) are not exact zero in the blue surface after linearization. In other words, the effect of $\phi$ and $\theta$ is not totally suppressed. While the coefficient of $\phi$ and $\theta$ are exact zero in the red surface after linearization; the contribution of $\phi$ and $\theta$ is totally cancelled.

Further thorough discussions on this phenomenon are beyond the scope of this paper.

### III. GENERALIZED TWO COLOR MAP THEOREM

Although Ref. [30] gives a complete Two Color Map, the designation of the robust gaits still follow the rectangular $(\alpha_1(t), \alpha_2(t))$ defined in (1) ~ (2). These requirements guarantee the continuity of the gaits. They may, on the other hand, rule out some robust gaits. E.g., the gait whose $\dot{\alpha}_1(t_1) \neq 0$ and $\dot{\alpha}_2(t_1) \neq 0$ at a time point $t_1$ is not included by Two Color Map Theorem since it violates (1).

With these concerns, we generalize Two Color Map Theorem in this section; Generalized Two Color Map Theorem gives more freedom to the gait plan on $\alpha_1 \in [-1,1] \cap \alpha_2 \in [-1,1]$.

## A. Initial $(\alpha_1(t), \alpha_2(t))$ and Color

The first step in planning a gait is determining the initial $(\alpha_1, \alpha_2)$ at $t = 0$ as well as the color of this initial $(\alpha_1, \alpha_2)$. Paint it either blue or red.

The first restriction of the initial $(\alpha_1, \alpha_2)$ is

$$\alpha_1 \in [-1,1] \cap \alpha_2 \in [-1,1]. \quad (3)$$

If the initial $(\alpha_1, \alpha_2)$ is painted in blue, the only restriction that it should obey is (3).

While if the initial $(\alpha_1, \alpha_2)$ is painted in red, it should further satisfy the following initial restrictions in (4), which is equivalent to (5).

$$(\alpha_1, \alpha_2) \notin \mathcal{L}, \quad (4)$$
$$(\alpha_1, \alpha_2) \in S_U \cup S_D, \quad (5)$$

where $\mathcal{L}$ is a curve within the region (3), defined by (6), space $S_U$ belongs to region (3), defined by (8), and space $S_D$ belongs to region (3), defined by (9), respectively.

$$\mathcal{L}: R(\alpha_1, \alpha_2) = 0, \quad (6)$$

where

$R(\alpha_1, \alpha_2) = 4.000 \cdot c1 \cdot c2 \cdot c1 \cdot c2 + 5.592 \cdot c1 \cdot c2 \cdot c1 \cdot s2 - 5.592 \cdot c1 \cdot c2 \cdot s1 \cdot c2 + 5.592 \cdot c1 \cdot s2 \cdot c1 \cdot c2 - 5.592 \cdot s1 \cdot c2 \cdot c1 \cdot c2 + 0.9716 \cdot c1 \cdot c2 \cdot s1 \cdot s2 + 0.9716 \cdot c1 \cdot s2 \cdot s1 \cdot c2 + 0.9716 \cdot s1 \cdot c2 \cdot c1 \cdot s2 + 0.9716 \cdot s1 \cdot s2 \cdot c1 \cdot c2 - 2.000 \cdot c1 \cdot s2 \cdot c1 \cdot s2 - 2.000 \cdot s1 \cdot c2 \cdot s1 \cdot c2 - 0.1687 \cdot c1 \cdot s2 \cdot s1 \cdot s2 + 0.1687 \cdot s1 \cdot c2 \cdot s1 \cdot s2 - 0.1687 \cdot s1 \cdot s2 \cdot c1 \cdot s2 + 0.1687 \cdot s1 \cdot s2 \cdot s1 \cdot c2,$

$$(7)$$

where $si = \sin(\alpha_i)$, $ci = \cos(\alpha_i)$, $(i = 1,2)$.

Fig. 6 visualizes $\mathcal{L}$ in $\alpha_1 - \alpha_2$ diagram. It is a black curve that divides the space (3) into 2 separate subspaces, one upper subspace above $\mathcal{L}$ and one lower space below $\mathcal{L}$.

It should be pointed out that a special point, Point $\mathcal{P}(\alpha_1, \alpha_2) = (0.167099, -0.167099)$, belongs to the lower subspace in Fig. 6. In other words, $\mathcal{P}$ lies below $\mathcal{L}$ rather than on it.

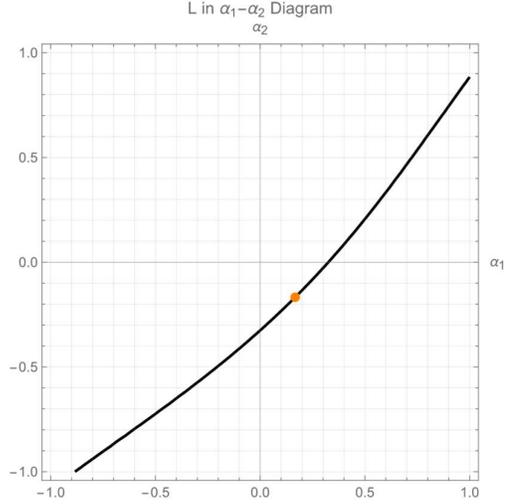

Fig. 6. Curve $\mathcal{L}$ in $\alpha_1 - \alpha_2$ diagram. It divides the whole space into two subspaces, one upper space above $\mathcal{L}$ and one lower space below $\mathcal{L}$. The orange point is at $(\alpha_1, \alpha_2) = (0.167099, -0.167099)$. Note that this point is a little below $\mathcal{L}$ rather than on $\mathcal{L}$.

$S_U$ and $S_D$ satisfy

$$S_U: R(\alpha_1, \alpha_2) > 0, \quad (8)$$
$$S_D: R(\alpha_1, \alpha_2) < 0. \quad (9)$$

Actually, $S_U$ is exactly the upper subspace in Fig. 6. While $S_D$ is exactly the lower subspace in Fig. 6. Obviously, $\mathcal{P} \in S_D$.

## B. Start $(\alpha_1(t), \alpha_2(t))$ in Red inside $S_U$

If $(\alpha_1(t), \alpha_2(t))$ starts in red inside $S_U$, $(\alpha_1(t), \alpha_2(t))$ is not allowed to escape from $S_U$ for any later time, e.g., $t > 0$.

Further, $(\alpha_1(t), \alpha_2(t))$ should be always red for any given time point while being continuous, e.g., satisfying (10).

$$\forall t_1 \geq 0, \lim_{t \to t_1^+} \alpha_i(t) = \lim_{t \to t_1^-} \alpha_i(t) = \alpha_i(t_1), i = 1,2. \quad (10)$$

This completes the Generalized Two Color Map Theorem for the case where $(\alpha_1(t), \alpha_2(t))$ starts in Red inside $S_U$.

## C. Start $(\alpha_1(t), \alpha_2(t))$ in Red inside $S_D$ or Start $(\alpha_1(t), \alpha_2(t))$ in Blue at Any Position

This section discusses the case of starting $(\alpha_1(t), \alpha_2(t))$ in red inside $S_D$ and the case of starting $(\alpha_1(t), \alpha_2(t))$ in blue at any position.

If the current color of $(\alpha_1(t), \alpha_2(t))$ is red inside $S_D$, $(\alpha_1(t), \alpha_2(t))$ is not allowed to escape from $S_D$ for any later time, e.g., $t > 0$, given that $(\alpha_1(t), \alpha_2(t))$ is always red for any subsequent time points.

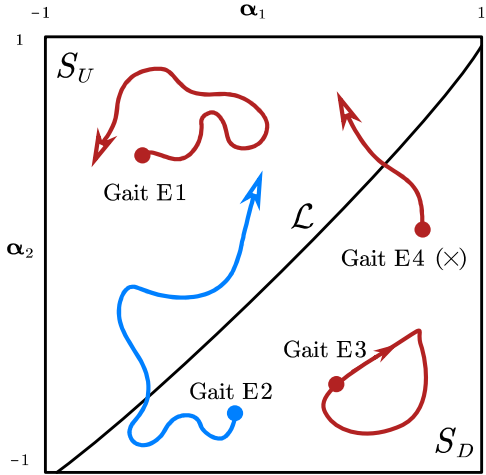

Fig. 7. Four examples of gaits. Gait E1, Gait E3, and Gait E4 start with red $(\alpha_1, \alpha_2)$. Gait E2 starts with blue $(\alpha_1, \alpha_2)$. Gait E1 follows Generalized Two Color Map Theorem in III. B. Gait E2 and Gait E3 follow Generalized Two Color Map Theorem in III. C. Gait E4 does not follow the Generalized Two Color Map Theorem in III. C; since it should not escape $S_D$ if its current color is red. So Gait E4 is not a robust gait.

If the current color of $(\alpha_1(t), \alpha_2(t))$ is blue at any position, $(\alpha_1(t), \alpha_2(t))$ is allowed to travel inside the entire space defined by (3) for any later time, e.g., $t > 0$, given that $(\alpha_1(t), \alpha_2(t))$ is always blue for any subsequent time points.

Still, $(\alpha_1(t), \alpha_2(t))$ is required to be continuous at any time, e.g., satisfying (10).

As an illustration, Fig. 7 gives four gaits, Gait E1, Gait E2, Gait E3, and Gait E4, on the $\alpha_1 - \alpha_2$ diagram. Gait E1 satisfies the requirements in III. B. Gait E2 and Gait E3 satisfy the requirements in III. C. So, Gait E1 ~ E3 are robust gaits satisfying Generalized Two Color Map Theorem. While Gait E4 violates the requirements in III. C. So, Gait E4 is not a robust gait.

Note that it is possible to change the color of $(\alpha_1, \alpha_2)$ at the special point, $\mathcal{P}$, on $\alpha_1(t) - \alpha_2(t)$ diagram. This color switch is detailed in III. D.

*D. Color Switch*

$(\alpha_1, \alpha_2)$ with the initial condition discussed in III. C is in the same color for the time points later. While there is a special point, on $\alpha_1(t) - \alpha_2(t)$ diagram, where changing the color of the current $(\alpha_1, \alpha_2)$ is possible.

At point $\mathcal{P}(\alpha_1, \alpha_2) = (0.167099, -0.167099)$, $(\alpha_1, \alpha_2)$ is able to change the color, e.g., from red to blue or from blue to red. Note that $(\alpha_1, \alpha_2)$ can change the color if and only if it passes this point; changing the current color at any other position is not allowed. Note that this color change is not obligatory; the same color can be chosen for $(\alpha_1, \alpha_2)$ after passing $\mathcal{P}$.

Note that after changing the color, $(\alpha_1, \alpha_2)$ should follow the rule in III. C based on the new current color. E.g., $(\alpha_1, \alpha_2)$ is allowed to travel in the whole space defined in (3) if its new current color is blue or is not allowed to escape $S_D$ if its new current color is red.

A gait with color change will be discussed in Section IV.

*E. Proof of Continuity*

Since we define $(\alpha_3, \alpha_4)$ in this research on the Two Color Map [29], the resulting gait $(\alpha_1, \alpha_2, \alpha_3, \alpha_4)$ is robust. This section proves the continuity of the resulting $(\alpha_3, \alpha_4)$.

Based on Ref. [29], the $(\alpha_3, \alpha_4)$ corresponding to red $(\alpha_1, \alpha_2)$ satisfy

$$\begin{cases} \alpha_3 = \alpha_1 \\ \alpha_4 = \alpha_2 \end{cases}. \quad (11)$$

And, the $(\alpha_3, \alpha_4)$ corresponding to blue $(\alpha_1, \alpha_2)$ satisfy

$$\begin{cases} \alpha_3 = -\alpha_1 + 0.334198 \\ \alpha_4 = -\alpha_2 - 0.334198 \end{cases}. \quad (12)$$

Obviously, $(\alpha_3, \alpha_4)$ is continuous corresponding to the $(\alpha_1, \alpha_2)$ of the same color, given that $(\alpha_1, \alpha_2)$ is changing continuously, e.g., satisfying (10). Thus, $(\alpha_1, \alpha_2, \alpha_3, \alpha_4)$ is a continuous gait for case with $(\alpha_1, \alpha_2)$ of same color.

Notice that Point $\mathcal{P}(\alpha_1, \alpha_2) = (0.167099, -0.167099)$ results the identical $(\alpha_3, \alpha_4)$ in both (11) and (12), enabling to change the color of $(\alpha_1, \alpha_2)$ while holding a continuous $(\alpha_3, \alpha_4)$, e.g., the left limit equals to the right limit, which equals to the value defined at Point $P$ for both $\alpha_1$ and $\alpha_2$. Thus, switching the color at $\mathcal{P}$ will not destroy the continuity of the gait. In conclusion, $(\alpha_3, \alpha_4)$ is always continuous.

Further, $(\alpha_1, \alpha_2)$ is continuous. Thus, the Generalized Two Color Map Theorem defines the continuous robust gaits, $(\alpha_1, \alpha_2, \alpha_3, \alpha_4)$.

The subspace-division and $\mathcal{L}$ guarantees that the resulting determinant of the decoupling matrix will be non-zero; it will either be always positive or be always negative. Crossing $\mathcal{L}$ while not following the requirements defined by Generalized Two Color Map Theorem will result a crossover at zero in that determinant of the decoupling matrix.

We refer the detail of this discussion to our previous publication [29].

## IV. ROBUSTNESS ANALYSIS

This section discusses the robustness of three periodic gaits, Gait 1, Gait 2, and Gait 3, that satisfy Generalized Two Color Map Theorem. Gait 1 undergoes color switch every half period. Gait 2 and 3 are changing in a circular pattern on $\alpha_1(t) - \alpha_2(t)$ diagram. Notice that all these three gaits does not satisfy Two Color Map Theorem since (1) ~ (2) are violated.

All these gaits are illustrated in Fig. 8.

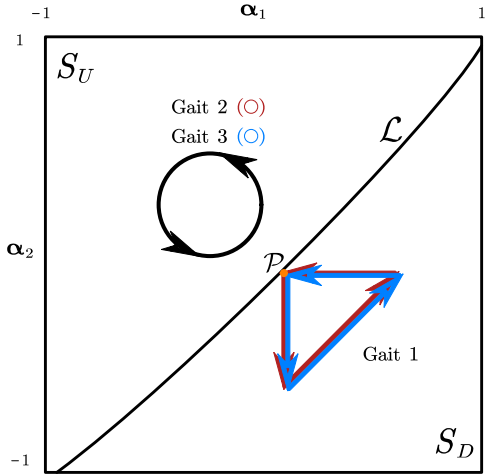

Fig. 8. The robustness of three periodic gaits is to be analyzed. They are Gait 1, Gait 2, and Gait 3. Gait 1 travels along a triangular pattern, starting in red on point $\mathcal{P}$. It changes its color every time it passes $\mathcal{P}$. Both Gait 2 and Gait 3 travel along a circular pattern, starting in red and blue, respectively. No color change happens in Gait 2 or Gait 3.

Both Gait 2 and Gait 3 travels along a circle on $\alpha_1(t) - \alpha_2(t)$ diagram. While Gait 2 is red and Gait 3 is blue, which determine $(\alpha_3, \alpha_4)$ by (11) and (12), respectively. Both gaits maintain their won color throughout the time; no color change happens to these gaits.

Gait 1 starts with red color on the switching point, $\mathcal{P}$. After half period, it returns to point $\mathcal{P}$ and changes its color to blue before the next half period as blue. After a whole period, it returns to $\mathcal{P}$ and changes it color to red to begin a new cycle.

In comparison, the corresponding adjacent biased gaits are created for each periodic gait by partially scaling, that is

$$\alpha_1 \leftarrow \alpha_1, \quad (13)$$
$$\alpha_2 \leftarrow \alpha_2, \quad (14)$$
$$\alpha_3 \leftarrow \eta \cdot \alpha_3, \quad (15)$$
$$\alpha_4 \leftarrow \eta \cdot \alpha_4, \quad (16)$$

where $\eta$ is the scaling coefficient. In this experiment, $\eta = 80\%$ for all the biased gaits.

### A. Gait 1

The angle history of Gait 1 is plotted in Fig. 9. The unacceptable attitudes for Gait 1 and the biased Gait 1 are illustrated in Fig. 10 in red curves and blue curves, respectively.

Clearly, the biased Gait 1 leaves little acceptable attitudes for the tilt-rotor. While the unacceptable attitudes for Gait 1 are relatively far from $(\phi, \theta) = (0,0)$. This indicates that Gait 1 is more robust to the biased Gait 1.

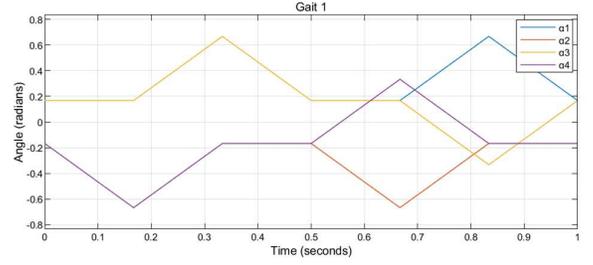

Fig. 9. The tilting angles in Gait 1 in the first period.

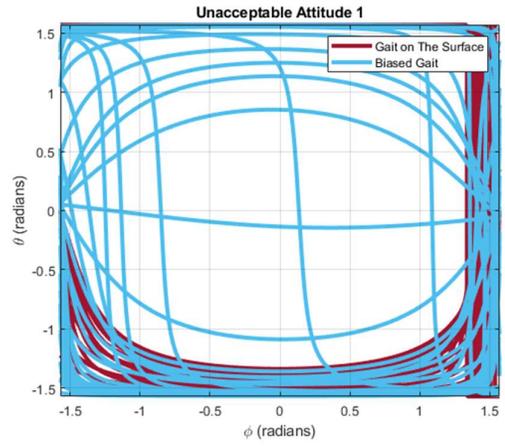

Fig. 10. The unacceptable attitudes for Gait 1 (red curves) and its biased gait (blue curves). The red curves represent the attitudes that will introduce the singular decoupling matrix in feedback linearization for Gait 1. The blue curves represent the attitudes that will introduce the singular decoupling matrix in feedback linearization for the biased Gait 1.

### B. Gait 2

The angle history of Gait 2 is plotted in Fig. 11. The unacceptable attitudes for Gait 2 and the biased Gait 2 are illustrated in Fig. 12 in red curves and blue curves, respectively.

Interestingly, though the biased Gait 2 leaves relatively abundant acceptable attitudes for the tilt-rotor, there are no unacceptable attitudes for Gait 2; any attitude in this figure are acceptable that will introduce an invertible decoupling matrix, given that the input is not saturated.

Thus, Gait 2 is more robust to the biased Gait 2.

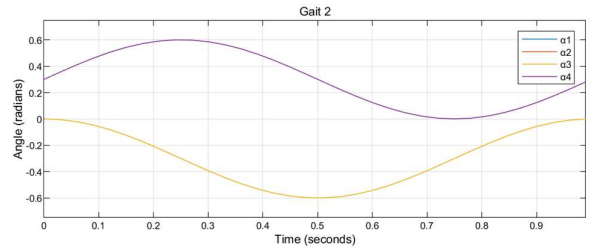

Fig. 11. The tilting angles in Gait 2 in the first period.

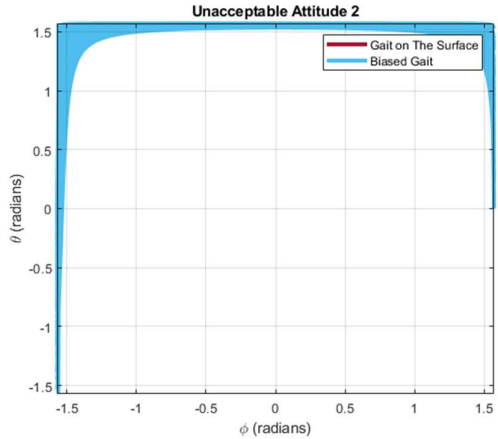

Fig. 12. The unacceptable attitudes for Gait 2 (red curves) and its biased gait (blue curves). The red curves represent the attitudes that will introduce the singular decoupling matrix in feedback linearization for Gait 2. The blue curves represent the attitudes that will introduce the singular decoupling matrix in feedback linearization for the biased Gait 2. Notice that there are no red curves in this diagram at all.

*C. Gait 3*

The angle history of Gait 3 is plotted in Fig. 13. The unacceptable attitudes for Gait 3 and the biased Gait 3 are illustrated in Fig. 14 in red curves and blue curves, respectively.

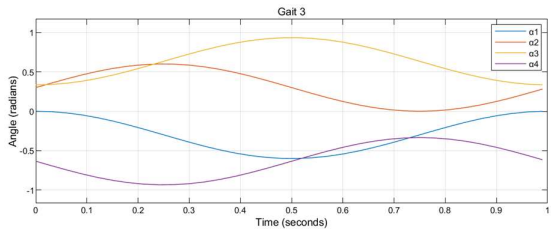

Fig. 13. The tilting angles in Gait 3 in the first period.

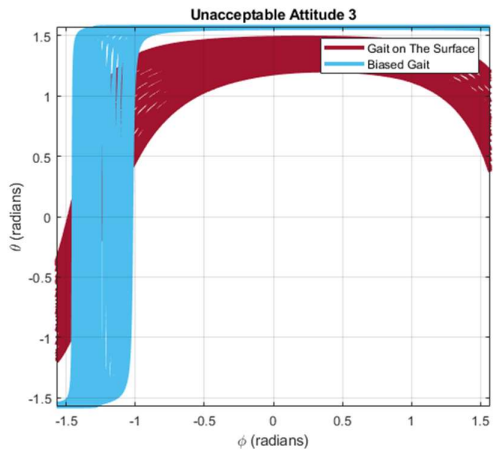

Fig. 14. The unacceptable attitudes for Gait 3 (red curves) and its biased gait (blue curves). The red curves represent the attitudes that will introduce the singular decoupling matrix in feedback linearization for Gait 3. The blue curves represent the attitudes that will introduce the singular decoupling matrix in feedback linearization for the biased Gait 3.

It is hard to assert whether Gait 3 or the biased Gait 3 is more robust. Indeed, both gaits are relatively far from $(\phi, \theta) = (0,0)$, leaving sufficient attitude space for the tilt-rotor.

While it is not surprising to receive this result – gaits on the Two Color Map Theorem or Generalized Two Color Map Theorem generates robust gaits, ensuring the attitudes near $(\phi, \theta) = (0,0)$ are acceptable for the tilt-rotor. The theorem loses its effects for the attitudes far from the origin.

## V. CONCLUSION

This research paper generalized the Two Color Map Theorem by relaxing the restrictions of the gait plan. With the sound proof deduced in this paper, the gaits planned following Generalized Two Color Map Theorem is always continuous and robust.

Comparing with the Two Color Map Theorem, the generalized theorem provides the robust gaits which are ignored by the previous former theorem. Three ignored gaits are evaluated and found their strong robustness to the attitude change. Besides, the robust gaits with color switch is evaluated for the first time in this paper.

In our future plan, we will develop tracking controller for the tilt-rotor adopting the robust gaits, relying on Generalized Two Color Map Theorem.

REFERENCES

[1] M. Ryll, H. H. Bulthoff, and P. R. Giordano, "Modeling and control of a quadrotor UAV with tilting propellers," in *2012 IEEE International Conference on Robotics and Automation*, St Paul, MN, USA: IEEE, May 2012, pp. 4606–4613. doi: 10.1109/ICRA.2012.6225129.

[2] Z. Shen and T. Tsuchiya, "Gait Analysis for a Tiltrotor: The Dynamic Invertible Gait," *Robotics*, vol. 11, no. 2, Art. no. 2, Apr. 2022, doi: 10.3390/robotics11020033.

[3] Z. Shen, Y. Ma, and T. Tsuchiya, "Quad-cone-rotor: a novel tilt quadrotor with severe-fault-tolerant ability," in *Unmanned Systems Technology XXIV*, P. L. Muench, H. G. Nguyen, and B. K. Skibba, Eds., Orlando, United States: SPIE, May 2022, p. 2. doi: 10.1117/12.2618468.

[4] M. Ryll, H. H. Bulthoff, and P. R. Giordano, "A Novel Overactuated Quadrotor Unmanned Aerial Vehicle: Modeling, Control, and Experimental Validation," *IEEE Trans. Contr. Syst. Technol.*, vol. 23, no. 2, pp. 540–556, Mar. 2015, doi: 10.1109/TCST.2014.2330999.

[5] M. Ryll, H. H. Bulthoff, and P. R. Giordano, "First flight tests for a quadrotor UAV with tilting propellers," in *2013 IEEE International Conference on Robotics and Automation*, Karlsruhe, Germany: IEEE, May 2013, pp. 295–302. doi: 10.1109/ICRA.2013.6630591.

[6] A. M. Ahmed and J. Katupitiya, "Modeling and Control of a Novel Vectored-Thrust Quadcopter," *Journal of Guidance, Control, and Dynamics*, vol. 44, no. 7, pp. 1399–1409, Jul. 2021, doi: 10.2514/1.G005467.

[7] A. Oosedo, S. Abiko, S. Narasaki, A. Kuno, A. Konno, and M. Uchiyama, "Flight control systems of a quad tilt rotor Unmanned Aerial Vehicle for a large attitude change," in *2015 IEEE International Conference on Robotics and Automation (ICRA)*, Seattle, WA, USA: IEEE, May 2015, pp. 2326–2331. doi: 10.1109/ICRA.2015.7139508.

[8] T. Magariyama and S. Abiko, "Seamless 90-Degree Attitude Transition Flight of a Quad Tilt-rotor UAV under Full Position Control," in *2020 IEEE/ASME International Conference on Advanced Intelligent Mechatronics (AIM)*, Boston, MA, USA: IEEE, Jul. 2020, pp. 839–844. doi: 10.1109/AIM43001.2020.9158965.

[9] S. Park *et al.*, "ODAR: Aerial Manipulation Platform Enabling Omnidirectional Wrench Generation," *IEEE/ASME Trans.*

*Mechatron.*, vol. 23, no. 4, pp. 1907–1918, Aug. 2018, doi: 10.1109/TMECH.2018.2848255.
[10] S. Jin *et al.*, "Back-stepping control design for an underwater robot with tilting thrusters," in *2015 International Conference on Advanced Robotics (ICAR)*, Istanbul, Turkey: IEEE, Jul. 2015, pp. 1–8. doi: 10.1109/ICAR.2015.7251425.
[11] J. Kadiyam and M. Santhakumar, "Design and Implementation of Backstepping Controller for Tilting Thruster Underwater Robot," p. 6.
[12] N. Phong Nguyen, W. Kim, and J. Moon, "Observer-Based Super-Twisting Sliding Mode Control with Fuzzy Variable Gains and its Application to Overactuated Quadrotors," in *2018 IEEE Conference on Decision and Control (CDC)*, Miami Beach, FL: IEEE, Dec. 2018, pp. 5993–5998. doi: 10.1109/CDC.2018.8619571.
[13] T. Blandino, A. Leonessa, D. Doyle, and J. Black, "Position Control of an Omni-Directional Aerial Vehicle for Simulating Free-Flyer In-Space Assembly Operations," in *ASCEND 2021*, Las Vegas, Nevada & Virtual: American Institute of Aeronautics and Astronautics, Nov. 2021. doi: 10.2514/6.2021-4100.
[14] D. Lu, C. Xiong, Z. Zeng, and L. Lian, "Adaptive Dynamic Surface Control for a Hybrid Aerial Underwater Vehicle With Parametric Dynamics and Uncertainties," *IEEE J. Oceanic Eng.*, vol. 45, no. 3, pp. 740–758, Jul. 2020, doi: 10.1109/JOE.2019.2903742.
[15] M. Hamandi, F. Usai, Q. Sablé, N. Staub, M. Tognon, and A. Franchi, "Design of multirotor aerial vehicles: A taxonomy based on input allocation," *The International Journal of Robotics Research*, vol. 40, no. 8–9, pp. 1015–1044, Aug. 2021, doi: 10.1177/02783649211025998.
[16] A. Nemati and M. Kumar, "Modeling and control of a single axis tilting quadcopter," in *2014 American Control Conference*, Portland, OR, USA: IEEE, Jun. 2014, pp. 3077–3082. doi: 10.1109/ACC.2014.6859328.
[17] Z. Shen and T. Tsuchiya, "A Novel Formula Calculating the Dynamic State Error and Its Application in UAV Tracking Control Problem," *arXiv:2108.07968 [cs, eess]*, Aug. 2021, Accessed: Jan. 10, 2022. [Online]. Available: http://arxiv.org/abs/2108.07968
[18] Z. Shen, "Flight Control System Design for Autonomous Aerial Surveys of Volcanoes," *arXiv:2111.11861 [cs, eess]*, Nov. 2021, Accessed: Jan. 10, 2022. [Online]. Available: http://arxiv.org/abs/2111.11861
[19] Z. Shen and T. Tsuchiya, "The Pareto-Frontier-Based Stiffness of a Controller: Trade-off Between Trajectory Plan and Controller Design," in *Intelligent Computing*, K. Arai, Ed., in Lecture Notes in Networks and Systems. Cham: Springer International Publishing, 2022, pp. 825–844. doi: 10.1007/978-3-031-10464-0_57.
[20] Z. Shen and T. Tsuchiya, "State Drift and Gait Plan in Feedback Linearization Control of A Tilt Vehicle," in *Computer Science & Information Technology (CS & IT)*, Vienna, Austria: Academy & Industry Research Collaboration Center (AIRCC), Mar. 2022, pp. 1–17. doi: 10.5121/csit.2022.120501.
[21] Z. Shen, Y. Ma, and T. Tsuchiya, "Stability Analysis of a Feedback-linearization-based Controller with Saturation: A Tilt Vehicle with the Penguin-inspired Gait Plan," *arXiv preprint arXiv:2111.14456*, 2021.
[22] R. Kumar, A. Nemati, M. Kumar, R. Sharma, K. Cohen, and F. Cazaurang, "Tilting-Rotor Quadcopter for Aggressive Flight Maneuvers Using Differential Flatness Based Flight Controller," presented at the ASME 2017 Dynamic Systems and Control Conference, Tysons, Virginia, USA: American Society of Mechanical Engineers, Oct. 2017, p. V003T39A006. doi: 10.1115/DSCC2017-5241.
[23] Z. Shen and T. Tsuchiya, "Cat-Inspired Gaits for a Tilt-Rotor—From Symmetrical to Asymmetrical," *Robotics*, vol. 11, no. 3, Art. no. 3, Jun. 2022, doi: 10.3390/robotics11030060.
[24] Z. Shen, Y. Ma, and T. Tsuchiya, "Feedback linearization-based tracking control of a tilt-rotor with cat-trot gait plan," *International Journal of Advanced Robotic Systems*, vol. 19, no. 4, p. 17298806221109360, Jul. 2022, doi: 10.1177/17298806221109360.
[25] J. A. Vilensky, J. Njock Libii, and A. M. Moore, "Trot-gallop gait transitions in quadrupeds," *Physiology & Behavior*, vol. 50, no. 4, pp. 835–842, Oct. 1991, doi: 10.1016/0031-9384(91)90026-K.
[26] Z. Shen and T. Tsuchiya, "Singular Zone in Quadrotor Yaw–Position Feedback Linearization," *Drones*, vol. 6, no. 4, Art. no. 4, Apr. 2022, doi: 10.3390/drones6040084.
[27] V. Mistler, A. Benallegue, and N. K. M'Sirdi, "Exact linearization and noninteracting control of a 4 rotors helicopter via dynamic feedback," in *Proceedings 10th IEEE International Workshop on Robot and Human Interactive Communication. ROMAN 2001 (Cat. No.01TH8591)*, Paris, France: IEEE, Sep. 2001, pp. 586–593. doi: 10.1109/ROMAN.2001.981968.
[28] J. Ghandour, S. Aberkane, and J.-C. Ponsart, "Feedback Linearization approach for Standard and Fault Tolerant control: Application to a Quadrotor UAV Testbed," *J. Phys.: Conf. Ser.*, vol. 570, no. 8, p. 082003, Dec. 2014, doi: 10.1088/1742-6596/570/8/082003.
[29] Z. Shen, Y. Ma, and T. Tsuchiya, "Four-Dimensional Gait Surfaces for a Tilt-Rotor—Two Color Map Theorem," *Drones*, vol. 6, no. 5, Art. no. 5, May 2022, doi: 10.3390/drones6050103.
[30] Z. Shen and T. Tsuchiya, "The Robust Gait of a Tilt-rotor and Its Application to Tracking Control -- Application of Two Color Map Theorem." arXiv, Jun. 26, 2022. doi: 10.48550/arXiv.2206.10941.
[31] I. Hameduddin and A. H. Bajodah, "Nonlinear generalised dynamic inversion for aircraft manoeuvring control," *International Journal of Control*, vol. 85, no. 4, pp. 437–450, Apr. 2012, doi: 10.1080/00207179.2012.656143.